\pdfoutput=1

\documentclass[11pt]{article}

\usepackage[]{EMNLP2022}

\usepackage{times}
\usepackage{algorithm}
\usepackage{algpseudocode}
\usepackage{xcolor}
\usepackage{latexsym}
\usepackage{graphicx}
\usepackage{booktabs}
\usepackage{multirow}
\usepackage{mwe}
\usepackage{subcaption}
\usepackage[T1]{fontenc}
\usepackage{hyperref}
\usepackage[utf8]{inputenc}

\usepackage{microtype}
\definecolor{conclusiongreen}{RGB}{1, 113, 0} 
\definecolor{reasonblue}{RGB}{0, 118, 186}

\title{Computing and Exploiting Document Structure to Improve Unsupervised Extractive Summarization of Legal Case Decisions}

\author{Yang Zhong\\
  University of Pittsburgh \\
 Pittsburgh, PA, USA \\
  \texttt{yaz118@pitt.edu} \\\And
 Diane Litman \\
  University of Pittsburgh \\
 Pittsburgh, PA, USA \\
  \texttt{dlitman@pitt.edu} \\}

\begin{document}
\maketitle
\begin{abstract}
Though many 
algorithms can be used to automatically summarize legal case decisions,  most 
fail to incorporate domain knowledge about how important sentences in a legal decision relate to a representation of its document structure. For example, analysis of a  legal case summarization dataset demonstrates that sentences serving different  types of argumentative roles in the decision appear in different sections 
of the document. 
In this work, we propose an unsupervised graph-based ranking model that uses a reweighting algorithm to exploit properties of the document structure of legal case decisions.  We also explore the impact of using different methods to compute the  document structure. Results on the Canadian Legal Case Law 
dataset  show that our proposed method outperforms several strong 
baselines. 


\end{abstract}
\section{Introduction}
Single document summarization aims at rephrasing a long text into a shorter version while preserving the important information \cite{INR-015}. While recent years have witnessed a blooming of abstractive summarization models that can generate fluent and coherent new wordings 
\cite{rush-etal-2015-neural, zhang2020pegasus, lewis-etal-2020-bart},  abstractive 
 summaries often contain hallucinated facts 
 \cite{kryscinski-etal-2019-neural}. In contrast, extractive summarization models directly select sentences/phrases from the source document to form a summary. In certain domains such as the law or science \cite{bhattacharya2019comparativestudy,dong-etal-2021-discourse}, using  exact wordings may be needed. 

In this work, we focus on {\it extractive summarization of legal case decisions}. Different from texts in the news domain, case texts tend to be longer  (e.g., in Canadian legal case decisions \cite{xu-2021-position-case} there are on average  3.9k words, while standard news articles \cite{nallapati-etal-2016-abstractive} range from 400 - 800 words) and also have more complicated document structures (e.g., legal cases are likely to be split into sections while news articles are not). 
In contrast to scientific domains, which also have long and structured texts, large 
training sets of case decisions and reference summaries are generally not freely  available given the restrictions of the legal field.  
A currently used case 
dataset has less than 30k training examples \cite{xu-2021-position-case}, which is ten times less than scientific datasets such as arXiv and PubMed \cite{cohan-etal-2018-discourse}. Thus, for the legal domain, it is not surprising that {\it unsupervised} extractive summarization methods are of 
interest. 
Unfortunately, when researchers \cite{sarvanan-2006-graphical, bhattacharya2019comparativestudy} have attempted to directly apply standard unsupervised 
models 
to legal data, they have obtained  mediocre results.  

\begin{figure}[t]
\centering
 \includegraphics[width=1.02\linewidth]{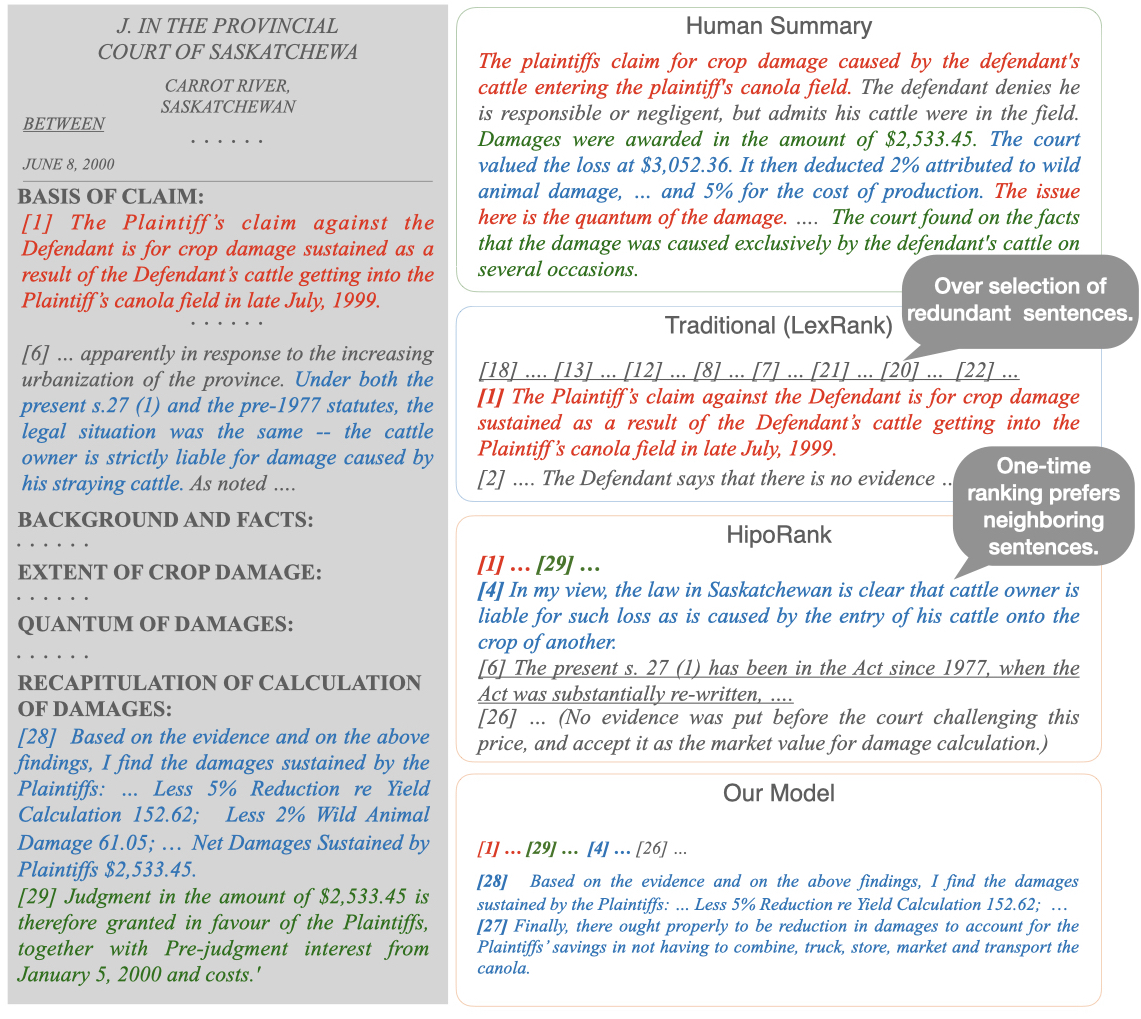}

  \caption{An example 
  case law document-summary pair (ID: 3\_2000canlii19612) and different summarization system outputs, where sentences are annotated with \textit{argumentative}  \textcolor{red}{Issue}, \textcolor{reasonblue}{Reason}, and \textcolor{conclusiongreen}{Conclusion} labels. Our 
  method better extracts  argumentative sentences from the source document by exploiting its structure.} 
  \label{fig:output_figure}
\end{figure}

However, most such attempts have failed to utilize the \textit{document structure
of legal texts}.
In case law, 
important sentences about the issues versus the decisions of the court occur in different places in the document structure.
In contrast,  summarization algorithms typically flatten any structure during initial processing (i.e., they concatenate sentences from different sections/paragraphs of a document to form a sentence list),
or select sentences using structural biases from other domains (e.g., the importance of leading sentences in news \cite{zheng-lapata-2019-sentence}).
As shown in Figure \ref{fig:output_figure}, while 
LexRank 
correctly extracts the legal issue from the beginning of the source text, it incorrectly extracts several redundant sentences (i.e., 
\textit{[20], [21]} and \textit{[22]} 
which talk about similar content) as well as ignores a large part of the article (e.g., no sentences are extracted from the last section of the original case: \textit{RECAPITULATION OF CALCULATION OF DAMAGES}). In contrast, the human  summary focuses on sentences related to the argument of the legal decision (e.g., what are the issues, reasoning and conclusions of this court case?), which tend to be spread across the document structure. 

Recently, the  HipoRank model was proposed to  exploit discourse structure patterns during unsupervised extractive summarization \cite{dong-etal-2021-discourse}.  However, the model was designed for  long scientific articles, and the experiments were based on data where the articles were already split into document sections.   
For case decisions,  document structures are generally either missing or only implicitly conveyed by text formatting.  For example, in Figure \ref{fig:output_figure}, document  sections are conveyed by bolding in the source HTML file. 
Moreover, algorithms such as PACSUM and HipoRank compute  sentence centrality just once and greedily select the top-k candidates 
as the extractive summary. 
Such a greedy {selection algorithm} fails to match the distribution of the argumentative sentences that ultimately appear in human case law summaries. 

To address these limitations, we investigate the utility of different methods for 
 automatically 
segmenting the sentences of legal case decisions into the sections of a document structure. We posit that incorporating {\it better views of  document structure} could bring improvements in summarization quality
when discourse-aware methods such as HipoRank are applied to legal case decisions.
We also propose a novel \textit{reweighting algorithm} to improve how HipoRank selects sentences when creating extractive summaries of legal decisions.  The algorithm
takes the history of already selected summary sentences into account, and gradually updates the importance score of a sentence.  We posit that reweighting will  decrease the selection of  redundant sentences as well as increase the selection of argumentative sentences from less-represented document segments (e.g.,  in the middle).

We evaluate our proposed method\footnote{Our code is available at \url{https://github.com/cs329yangzhong/DocumentStructureLegalSum}} for summarizing legal decisions using an annotated  Canadian case summarization dataset (CanLII) \cite{xu-2021-position-case}. Based on the belief that \textit{argumentative sentences} will capture the important sentences to summarize in a legal decision \cite{xu-2021-position-case,elaraby-litman-2022-arglegalsumm}, a portion of the CanLLI dataset comes with gold-standard sentence-level labels identifying which sentences are related to the issue/reasoning/conclusion of the court's decision in both source and summary documents. We use these labels to additionally propose a  metric that can better evaluate the quality of the generated summary from a legal expert's perspective. 
Empirical results show that our method
improves performance over previous unsupervised
models \cite{zheng-lapata-2019-sentence, dong-etal-2021-discourse, erkan2004lexrank} in automatic evaluation. 

\section{Related Work}


{\bf Supervised Extractive Summarization Using Discourse Information}
Graph-based methods have been exploited for extractive summarization tasks to better model the inter-sentence relations based on document structure. \citet{xu-etal-2020-discourse} applied a GCN layer to aggregate information from the document's discourse graph based on RST trees and dependencies.  More recently, 
 HiStruct+ \cite{ruan-etal-2022-histruct} and HEGEL \cite{zhang-2022-hegel} started to incorporate the hierarchical structure and topic structure of scientific articles into supervised model training, respectively. However, HiSruct+ relied on the relatively fixed and explicit document structure of scientific articles\footnote{Section titles following a shared pattern (introduction, method … and conclusion) are encoded to provide structural information. However, in our dataset, sectioning is often missing 
 or not meaningful (e.g.,  titles such as "section 1").},
 while HEGEL relied on a large training set to identify the topic distributions. \textit{Our work uses an unsupervised extractive summarization approach in a lower-resource setting, as well as studies the effects of computing different types of document structures.}
 We leave the exploration of the aforementioned supervised approaches on legal domain texts for future work.  

{\bf Unsupervised Extractive Summarization}
Traditional extractive summarization methods are mostly unsupervised \cite{radev-etal-2000-centroid, 10.5555/2832415.2832442,  hirao-etal-2013-single}, where a large portion apply the graph-based algorithms \cite{SALTON1997193, Steinberger2004UsingLS,erkan2004lexrank} or are based on term frequencies such as n-gram overlaps \cite{Nenkova2005TheIO} to rank the sentences' importance. More recently, pretrained transformer-based models \cite{devlin-etal-2019-bert, lewis-etal-2020-bart, zhang2020pegasus} have provided better sentence representations. 
For instance, \citet{zheng-lapata-2019-sentence} built directed unsupervised graph-based models on news articles using BERT-based sentence representations and achieved comparable performance to supervised models on multiple benchmarks. \citet{dong-etal-2021-discourse} 
augmented the document graph of \citet{zheng-lapata-2019-sentence}
with sentence position and section hierarchy to reflect the
document structure of scientific articles. \textit{Different from these two works which are based on assumptions of news and scientific article structures, our method uses reweighting to better utilize the document structure of legal cases.}

{\bf Extractive Summarization of Legal Texts}
Despite the success of supervised neural network  models in news and scientific article summarization \cite{zhang2020pegasus, lewis-etal-2020-bart,zaheer2020bigbird}, they face challenges in legal document summarization given the longer texts, distinct document structure, and limited training data \cite{bhattacharya2019comparativestudy}. Instead, prior work has tackled legal extractive summarization by applying domain independent unsupervised algorithms \cite{Luhn-1958-automatic,erkan2004lexrank,  sarvanan-2006-graphical}, or designing domain specific supervised approaches \cite{sarvanan-2006-graphical, polsley-etal-2016-casesummarizer,zhong-2019-iterativemasking}. One recent work \cite{bhattacharya2021incorporating} frames the task as Integer Linear Programming and demonstrates the importance of in-domain structure and legal knowledge. 
In another line of research, \citet{xu-2021-position-case} propose a sentence classification task with the hope of exploiting the court decision's \textit{argument structure} by making explicit its issues, conclusions, and reasons (i.e., argument triples). \textit{Our work is unsupervised
and implicitly reveals the relations between argument triples to generate better summaries.}
\section{Case Decision Summarization Dataset}\label{sec:dataset}
Recent work has introduced a number of legal document summarization or salient information identification tasks with associated datasets, e.g., for bill 
summarization 
\cite{kornilova-eidelman-2019-billsum}
and for case sentence argumentive classification \cite{xu-2021-position-case} and rhetorical role prediction
\cite{malik2021semantic}.
\begin{table}[t!]
    \centering
    \renewcommand{\arraystretch}{1}
    \begin{tabular}{l|c}
    
   \toprule
    Case length (avg. \# words) & 3,971\\
    Summary length  (avg. \# words) & 266\\
    Training set  (\# case/summary pairs) & 27,241 \\
    
    Testing set  (\# case/summary pairs) & 1,049\\
    
    \bottomrule
    \end{tabular}
    \caption{Dataset statistics of CanLII.}
    \label{tab:data_stats}
\end{table}
\begin{figure}[t!]
\centering
 \includegraphics[width=.9\linewidth]{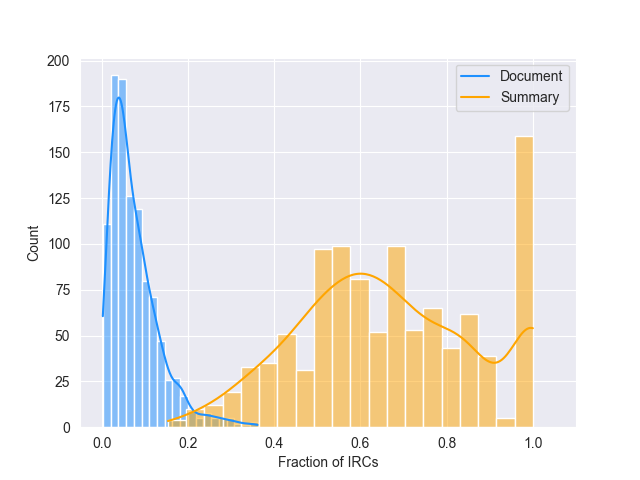}

  \caption{Fraction of sentences annotated as argumentative (using the IRC scheme) in the case documents versus in the summaries of the CanLII test set. Though only a small fraction of sentences in the original document are annotated as IRCs, IRCs are a large fraction of the human-written summaries.}
  \label{fig: disbribution}
\end{figure} 
Similarly to \citet{xu-2021-position-case}, we use the {\bf CanLII} (Canadian Legal Information Institute) dataset of legal case decisions and summaries\footnote{The data was obtained through an agreement with the Canadian Legal Information Institute (CanLII) (\url{https://www.canlii.org/en/}).}. 
  Full corpus statistics are provided in Table \ref{tab:data_stats}, while an example case/summary pair from the test set is provided in Figure \ref{fig:IRC_example} in Appendix \ref{appendix:1}.

\citet{xu-2021-position-case} only used a small portion of this dataset for their work in argumentative classification. Conjecturing that explicitly identifying the decision's argumentative components would be crucial in case summarization,  they annotated 1,049  case and human-written summary pairs curated from the full dataset.
In particular, they recruited legal experts to annotate the document on the sentence level, adopting an \textbf{IRC scheme} (see Figure \ref{fig:IRC_example} in Appendix \ref{appendix:1}) which classifies individual sentences into one of  four categories: \textbf{Issue} (legal question  addressed in the case), \textbf{Conclusion} (court’s decisions for the corresponding issue), \textbf{Reason} (text snippets explaining why the court made such conclusion) and \textbf{Non\_IRC} (none of the above). The distributions of the IRC labels in  the cases and summaries are shown in Figure \ref{fig: disbribution} and illustrate that argumentative sentences do indeed play an important role in human summaries. We utilized the unannotated 27,241  pairs to train a  supervised model baseline and the 1049 annotated pairs as our test set.  While none of our summarization methods  use the IRC annotations, they are used during testing as the basis of a domain-specific evaluation metric.


    

\section{Method and Models}
We propose a reweighting model that employs a 
graph-based ranking algorithm to exploit the 
structures encoded in long legal case decisions. 

\subsection{Discourse-Aware Backbone Model}\label{sec:main_paper_hiporank}
The HipoRank (Hierarchical and Positional Ranking) model recently developed by \citet{dong-etal-2021-discourse} constructs a directed graph for document representation using document section and sentence hierarchies.
HipoRank computes the centrality score of each sentence as 
\begin{equation}
    c(s_i^{I}) = \mu_1 c_{inter} (s_i^{I}) + c_{intra}(s_i^{I})
\end{equation} where $s_i^{I}$ refers to the $i$-th sentence in $I$-th section. $\mu_1$ is a tunable hyper-paramter, $c_{inter}(s_i^{I})$ computes the sentence's similarity to other section representations and  $c_{intra}(s_i^{I})$ computes the average similarity of the current sentence with all others in the same section. 
HipoRank then selects the top-K ranked sentences as the  summary. More details of the algorithm are provided in Appendix \ref{sec:hiporank}.
Directly applying HipoRank to our data yielded multiple challenges (e.g.,  redundant neighboring sentences 
(recall Figure \ref{fig:output_figure}) as well as too many sentences from the ends of the article were selected).

\subsection{Multiple Views of Document Structure}\label{sec:views}

\begin{figure*}[t!]
\centering
 \includegraphics[width=0.85\linewidth]{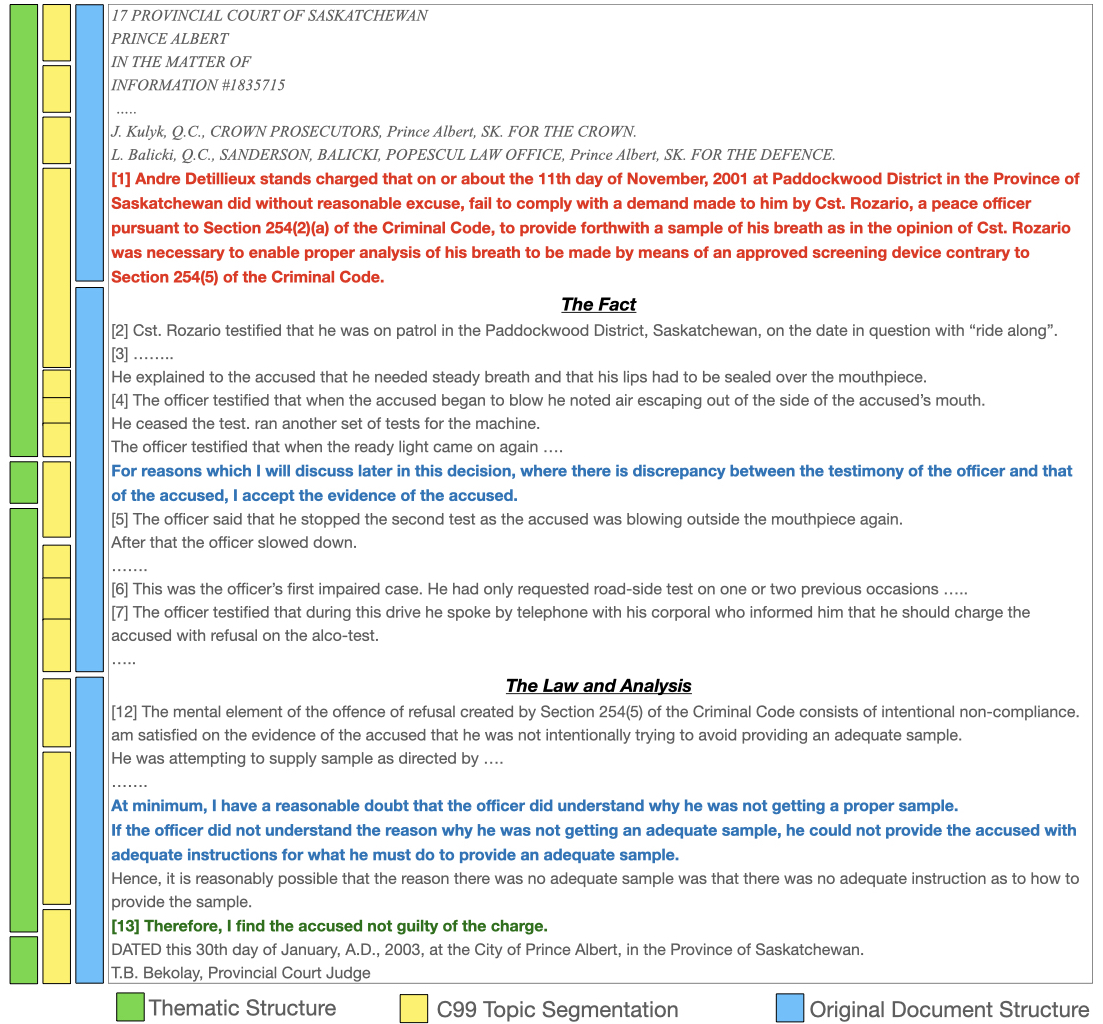}
  \caption{Different document structure views of a legal case decision (ID: c\_2003skpc17) from our CANLLI test set. Original case sentences are annotated with \textcolor{red}{Issue}, \textcolor{reasonblue}{Reason}, and \textcolor{conclusiongreen}{Conclusion} labels. On the left side, the \textcolor{green}{green}, \textcolor{yellow}{yellow} and \textcolor{cyan}{blue} boxes refer to thematic stage, topic segmentation and the original document structure, respectively. The boxes mean the corresponding sentences on the right hand side are grouped into the same segments. For instance, for the first blue box, the original article is split by the italicized and bolded section name of ``The Fact''. }
  \label{fig:three_view}
\end{figure*}

Before creating a HipoRank 
document graph, 
the document must be split into sections and sentences. 
The scientific 
datasets previously used with HipoRank 
were already split \cite{dong-etal-2021-discourse}.   We investigate {\it the summarization impact of using different approaches to automatically compute linear sections of the document structure}.
Figure~\ref{fig:three_view} shows different structures for the same  case.

\textbf{Original Document Structure}
This approach extracts the structure by processing the 
{\it HTML files}. We use a heuristic to mark the section names with an italic and bold format as the 
boundaries and segment the documents into multiple continuous sections. It is worth noting that 297 of the 1049 test case documents do not come with explicit section splits, thus we treat them as whole text spans\footnote{The Original Structure method processed HTML source files and split sentences using a legal 
sentence splitter (\url{https://github.com/jsavelka/luima_sbd}). The Topic and Thematic views used non-HTML data preprocessed by \citet{xu-2021-position-case}, but used the same sentence splitter.}.

\textbf{Topic Segment View} Meanwhile, we also explore using a traditional, {\it domain-independent linear text segmentation} algorithm. We use  C99 \cite{choi-2000-advances} but with advanced sentence representation from SBERT \cite{reimers-gurevych-2019-sentence} to group neighboring sentences into topic blocks. 

\textbf{Thematic Stage View}
Early studies found that legal documents tend to have well-defined, 
{\it domain-dependent thematic structures} \cite{farzindar-lapalme-2004-legal} or rhetorical roles \cite{saravanan-etal-2008-automatic}. Following work that extracts stage views in conversations (introductions → problem exploration →
problem solving → wrap up) \cite{chen-yang-2020-multi}, we extract thematic stages
through a Hidden Markov Model (HMM). A fixed order of stages is imposed and only consecutive transitions are allowed between neighboring stages.  We again represent the sentences 
using SBERT \cite{reimers-gurevych-2019-sentence} and set the number of stages as 5, including the starting Decision Data, Introduction, Context, Judicial Analysis, and Conclusion, as introduced by \citet{farzindar-lapalme-2004-legal}. 
 
\textbf{CanLII Header Removing}\label{sec:header_removing} Preliminary analysis demonstrated that the raw CanLII documents fail to distinguish the less important headers at the beginning  (i.e., the descriptive text before the main  content, for example,  the content above the grey splitting line and BASIS OF CLAIM in Figure \ref{fig:output_figure}). Generated summaries also tend to cover  a large portion of such information. We thus propose a legal case decision preprocessing procedure following certain heuristics\footnote{See Appendix \ref{sec:appendix_heuristics} for details.} to remove those headers, and demonstrate the improved summarization quality (for all views of document structure) in Section \ref{sec:result}. 




\subsection{Reweighting the Centrality Score}\label{sec:dynamic_selection}

The HipoRank document graph 
will not change once built, and the important sentences are greedily selected based on the aggregated centrality scores. 
This may introduce redundancies in selecting similar sentences and ignore the contents in the middle of the source case decisions that are more important once the argumentative sentences at the beginning and end are taken into account. We propose a novel reweighting approach that can tackle this problem. A prior attempt \cite{tao2021unsupervised}  on multi-round selection looked at the local similarity between 
selected sentences. They iteratively recompute the sentence to sentence similarities between the selected summary sentences and recompute the final sentence centrality scores after each sentence selection. Instead, we are focusing on modeling  the relationship between the selected sentence and the other candidate sentences.  Their method is also not directly applicable to longer text due to the $n^2$ time complexity of computation given large numbers of sentences (on average 205 sentences for CanLII dataset). 
\begin{algorithm}
\caption{Reweighting Algorithm}\label{alg:cap}
\begin{algorithmic}
\Require computed centrality score $c(s_i^{I})$ for all sentence s, $c_{intra}(d)$ for different section d 's embedding, and a threshold \textit{g} for phase transition and maximum summary length $max_{len}$.

\State $Summ \gets []$
\State \textbf{PHASE 1} 
\While{$len(Summ) \leq g * max_{len}$}

    \State $Summ.append(topK(\{c(s_i^{I})\})$
\EndWhile \\
\State{\textbf{PHASE 2}} 
\While{$len(Summ) \leq max_{len}$}
\State $c(s_i^{I}) \gets c(s_i^{I}) - sim(c_{intra}(I), c_{intra}(J)) * \mu_2$ 
\Comment{J is the section index of last selected sentence, $\mu_2$ is a hyperparameter}
\State $Summ.append(top-1({c(s_i^{I})}))$
\EndWhile

\textbf{Return} Summ
\end{algorithmic}
\label{algorithm_1}
\end{algorithm}

Our approach can be divided into two phases, as shown in Algorithm \ref{algorithm_1}. In the first phase, we assume that important argumentative sentences at the two ends of the document can be easily detected (as shown in Figures \ref{fig:output_figure} and \ref{fig:three_view}, legal case documents generally start by introducing the issues and end with  conclusions).  A quantitative analysis of the top-5 selected sentences in CanLII in fact provides an 80\% coverage of issue or conclusion sentences.   We thus set up a threshold to pick the first k sentences based on the original document graphs. Afterward, we gradually downweight the sentence's centrality score using the location of the latest selected sentence, that is, we set a penalty score for sentences that are placed as a neighbor of the current sentence selected for the summary. Our rationale 
is that 
reasoning sentences are more likely to be located in different sections in the middle that are not shared with issues and conclusions. 

\section{Experimental Setup}
For supervised models, we split the training data in an 80:20 ratio for training and validation. For unsupervised models, we tune the hyperparameters on the validation set. Model training details can be found in Appendix \ref{appendix:detail}. 



\begin{table*}[t]
    \centering
    \renewcommand{\arraystretch}{1}
    \begin{tabular}{c|l|cccc}
    \toprule
      ID & Model  &  R-1 & R-2 & R-L   & BS\\
         \midrule
         \multicolumn{6}{c}{{Oracles}} \\
         \midrule
          1 & IRC & 58.04 & 36.02 &55.28 & 87.94  \\
        2 &  EXT (ROUGE-L, F1) & 59.38 & 38.77 & 56.94 & 87.85 \\

    \midrule 
    \multicolumn{6}{c}{Extractive baselines (no document structure)} \\
    \midrule
    \multicolumn{6}{l}{\textit{supervised}} \\
    3 & BERT\_EXT & \textit{43.44} & {17.84} & \textit{40.36} & {84.47}  \\
   
        \multicolumn{6}{l}{\textit{unsupervised}} \\
       4 & LSA & 37.22 & 17.82 & 34.87 & {\bf \underline{\textit{84.48}}} \\
           5 & LexRank & 37.90 &\underline{\textit{18.17}} & 35.62 &  84.32 \\
            6 & TextRank  & 36.70 &16.19 &34.00 & 83.51 \\
            7 & PACSUM & 40.01 & 15.68 &37.37 & 83.52  \\
           \midrule 
           \multicolumn{6}{c}{{HipoRank backbone (with different computed document structures) 
           }} \\
          \midrule
          8 & Original Structure &  {41.61} & 17.13 & {38.73}  & 83.55  \\
          9 & C99-topic & 41.33 & 16.48 & 38.45 &  83.53  \\
          10  & HMM-stage & 40.71 & 15.64 & 37.93 & 83.57 \\
          11 & Original Structure w/o header & 42.58 & 18.01 & 39.63 & 83.62 \\
          12 & C99-topic  w/o header & \underline{43.25} & 18.02 & \underline{40.25} & \underline{\textit{\textbf{84.48}}} \\
        
          13 & HMM-stage  w/o header & 42.64 & 17.38 & 39.76 & 83.57\\
          
          \midrule 
              \multicolumn{6}{c}{{Ours (HipoRank backbone + Reweighting Algorithm)}} \\
              \midrule
          14 & Original Structure w/o header & 43.14 & {18.46} & 40.23 &  84.20  \\ 
          15 & C99-topic w/o header & \textbf{43.90} & \textbf{18.67} & \textbf{41.00} &  {84.34} \\
            16 & HMM-stage w/o header & {43.28} & 17.80 &40.40 &  84.22 \\ 
            
        \midrule
    \bottomrule
    \end{tabular}
    \caption{The automatic evaluation results on the CanLII test set. \textbf{Bold} represents the best non-oracle score, \textit{italic} the best baseline/backbone score, and \underline{underline} the best unsupervised baseline/backbone score.} 
    \label{tab:unsupervised_results}
\end{table*}

\textbf{Upper Bound Oracles}\label{sec:oracle}  
Based on  Figure \ref{fig: disbribution}, we create a domain-dependent \textbf{IRC\_Oracle} model where test sentences  manually annotated with the IRC  labels are concatenated to form the summary.   Following \citet{nallapati2017summarunner}, we also report results for \textbf{EXT\_Oracle}, a domain-independent
summarizer 
which greedily selects sentences from the original document based on the ROUGE-L scores compared to the abstractive human summary. 

\textbf{Extractive Baselines} 
For unsupervised models, we compare with LSA  \cite{Steinberger2004UsingLS},
LexRank \cite{erkan2004lexrank}, TextRank \cite{DBLP:journals/corr/BarriosLAW16}, and PACSUM \cite{zheng-lapata-2019-sentence}. We also include HipoRank \cite{dong-etal-2021-discourse} with document views. 
For supervised methods, we compare with BERT\_EXT \cite{liu-lapata-2019-text}. 
Although not our focus,  abstractive baselines are in Appendix \ref{sec:appendix_abstractive}. 

\textbf{Automic Evaluation Metrics}
We report ROUGE-1 (R-1), ROUGE-2 (R-2), and ROUGE-L (R-L) F1 scores, as well as 
BERTScore (BS) \cite{bert-score}. 
We also propose metrics to measure the recall value of the annotated IRC types in the test set, 
which exploits the structure of case documents. More details are in Section \ref{sec:salient_eval}.

\section{Results}\label{sec:result}
In this section, we aim to deal with three research questions: 
\textbf{RQ1}. How well do the extractive baselines including the HipoRank backbone deal with legal documents?  
\textbf{RQ2}. How well do the different views of document structures perform with the  HipoRank backbone?  \textbf{RQ3}. Can the reweighting algorithm help select important argumentative sentences and improve summary quality?

\subsection{Automatic Summarization Evaluation}
Table \ref{tab:unsupervised_results}
compares our methods with prior extractive models. See Appendix \ref{appendix:output} for example summaries. 


\textbf{RQ1.} Table \ref{tab:unsupervised_results} shows that there is still a gap between oracle models (rows 1 and 2) and current extractive baselines. There are around 20 points differences on R-1, R-2, and R-L. Among the baselines, the supervised model works best  only for R-1 and R-L. Unsupervised methods obtain  the highest BS (row 4) and R-2  (row 5), possibly due to the higher coverage of n-grams benefitting from longer extracted summaries (row 3 model generated summaries have an average length of 250; row 4-6 models generate on average 400-word summaries;  row 7 and 8 models have a limit of 220 words). 
Without  reweighting, the HipoRank backbone 
never outperforms the best extractive baseline.  However, if only unsupervised baselines are considered,  HipoRank in row 12 does show the best performance for 3 of the 4 evaluation metrics.

\textbf{RQ2.} To examine the effects of the  document views in  Section \ref{sec:views}, we split the document into different types of linear segments and then used  HipoRank  to generate summaries.  Recall that HipoRank is the only model to exploit  document structure, and as noted for RQ1, with the right structure could obtain the  best unsupervised R-1, R-L, and BS baseline scores.  
When naively constructing different document structures from the CanLII dataset without header removal, using NLP algorithms (rows 9 - 10) versus just using the HTML formatting (row 8) generally degraded results. However, when we  experimented with a modified version of the input documents (rows 11-13) where the headers were filtered through heuristics before computing the document structure, 
the  scores in rows 11-13 were higher (or in one case the same) than the comparable scores in rows 8-10.  
Also, without headers, the C99  topic segmentation algorithm (row 12) now outperforms the use of HTML (row 11) (obtaining an average improvement of 0.5 points across ROUGE and 0.8 for BS), suggesting
 that better  structures can  improve summarization. 
 As shown in Table \ref{tab:segmentation_stats} (and earlier in Figure \ref{fig:three_view}), C99  creates many small sections (average number of sentences per section is 3.39 with standard deviation  of 0.67).
 We hypothesize that this encourages the selection of sentences from more fine-grained segments. In contrast, the other two methods create lengthy sections (average of more than 50 sentences) with a large standard deviation (135.40 for original structure without headers). 
In sum, with improvements in automatic metrics, we find that document structures play an important role in summarizing cases.
\begin{table}[t]
\small
    \centering
    \begin{tabular}{c|c|c}
    \toprule
        Model & avg. \# secs & avg. \# sents per sec  \\
        \midrule 
        \multicolumn{3}{c}{with header} \\
        \midrule
        Original Structure & 4.83 (6.44) &  83.82 (118.78) \\
        C99-topic & 63.47 (70.34) & 3.38 (0.71) \\
        HMM-stage & 4.00(0.83) &  54.32 (64.80) \\
        \midrule
         \multicolumn{3}{c}{without header} \\
         \midrule
        Original Structure & 3.67 (5.51) &  102.99 (135.40) \\
        C99-topic & 59.74 (69.91) & 3.39 (0.67) \\
        HMM-stage & 3.16 (1.08) &  70.19 (119.39)\\
        \bottomrule
    \end{tabular}
    \caption{Statistics about the average number of sections (avg. \# secs) and average number of sentences per section (avg. \# sents per sec) across the documents with different computed document structures (standard deviation in parenthesis).}
    \label{tab:segmentation_stats}
\end{table}

\textbf{RQ3}. The final block of Table \ref{tab:unsupervised_results} presents the reweighting  results (using the "w/o header" version of the CanLLI documents as they performed best in the prior block).  
By downweighting  sentences that appear under the same section as  previously selected ones, we observe an F1 improvement of 0.65, 0.65, and 0.75 on R-1, R-2, and R-L, respectively, on the previously best-performing topic segmented document (row 12 versus 15).
Row 15 in fact has the best non-oracle results for all ROUGE scores.  This observation regarding the value of reweighting also holds for the original structure (row 11 vs. 14) and the HMM-stage segments (row 13 vs. 16). 



\begin{figure*}[h]
        \centering
        \begin{subfigure}[b]{0.45\textwidth}
            \centering
            \includegraphics[width=\textwidth]{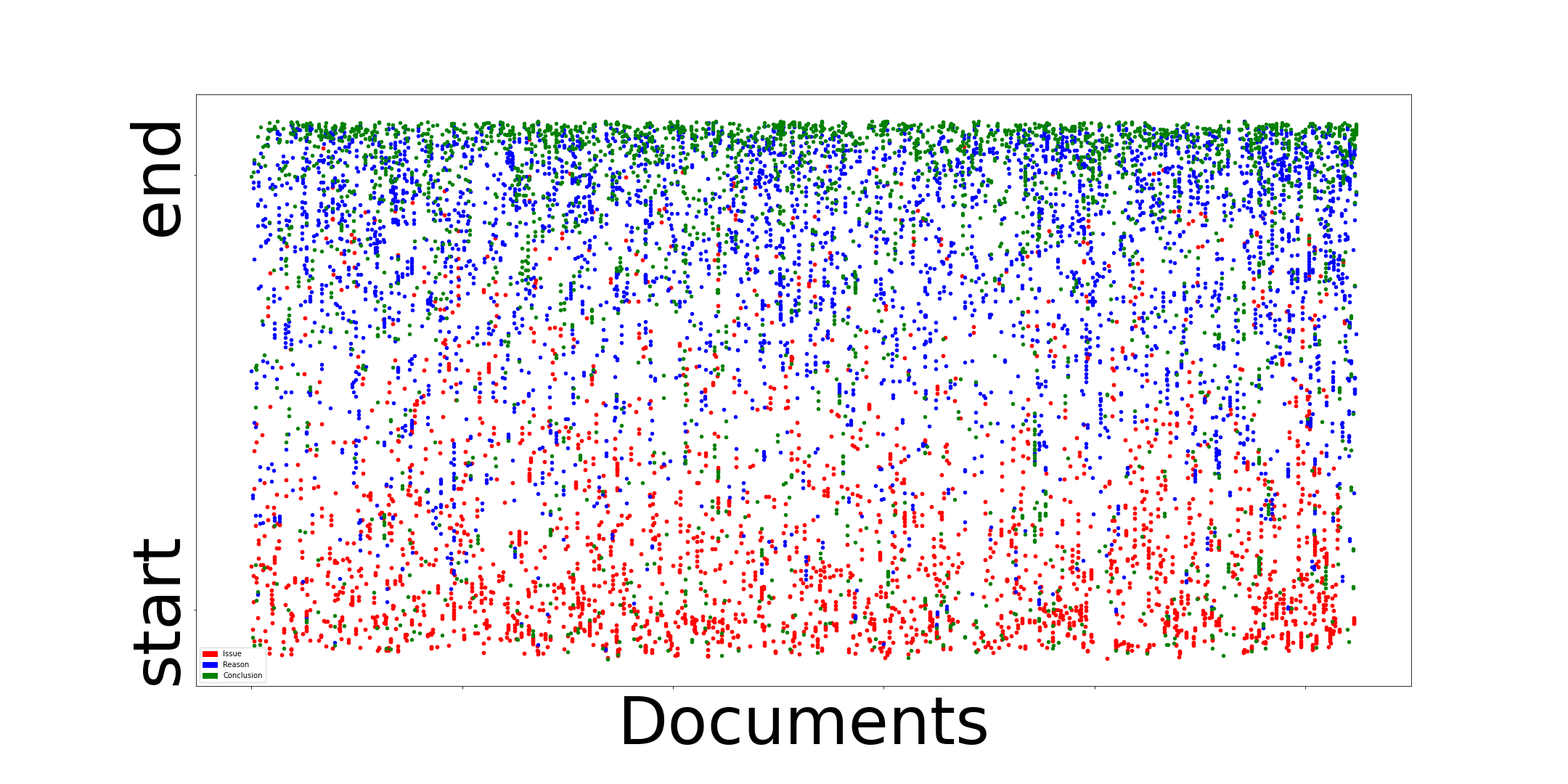}
            \caption[]%
            {{\small IRC Oracles}}    
            \label{IRCoracle}
        \end{subfigure}
        \hfill
        \begin{subfigure}[b]{0.45\textwidth}   
            \centering 
            \includegraphics[width=\textwidth]{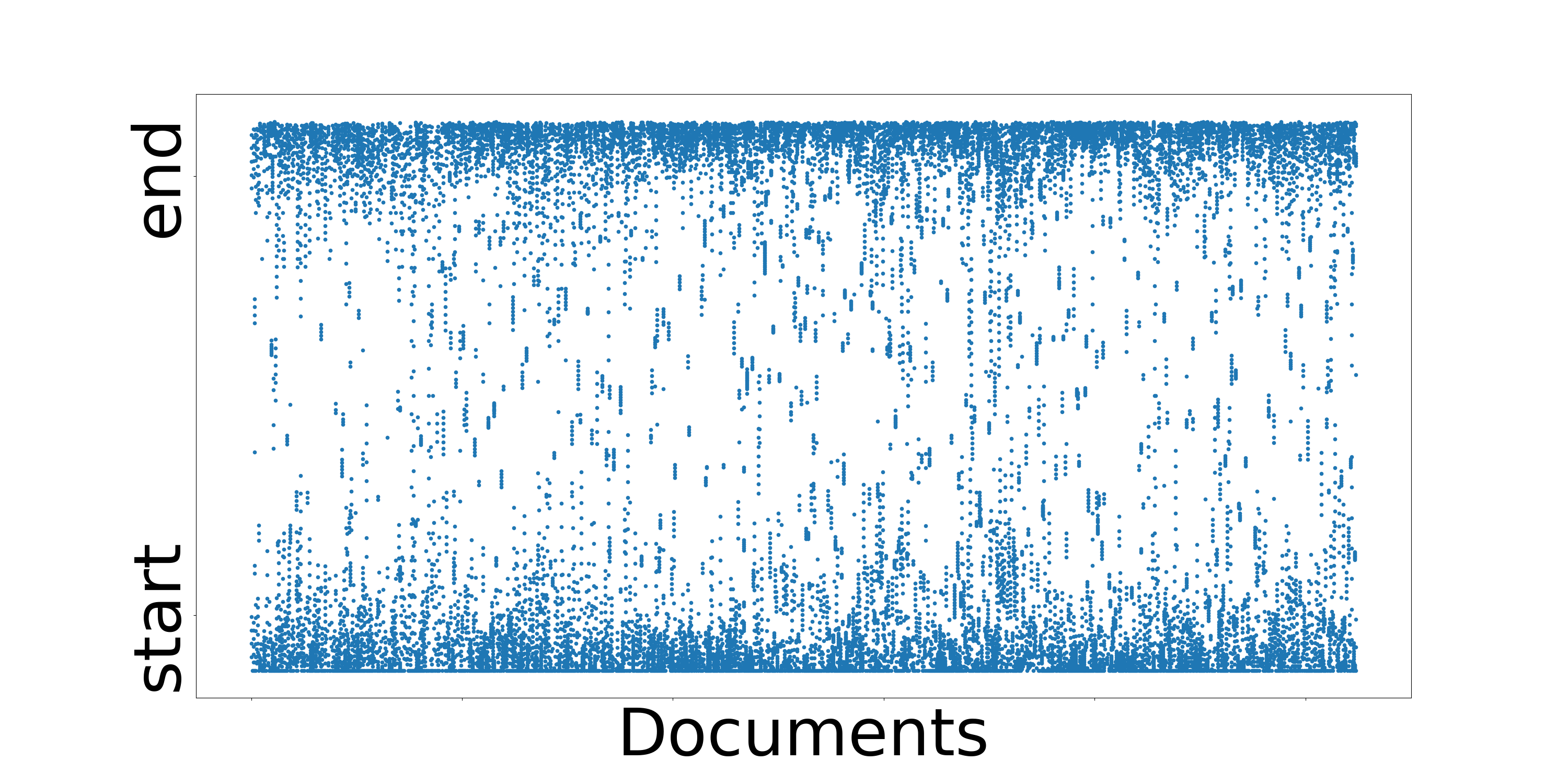}
            \caption[]%
            {{\small HipoRank with headers}}    
            \label{fig:mean and std of net34}
        \end{subfigure}
        \vskip\baselineskip
        
        \begin{subfigure}[b]{0.45\textwidth}
            \centering
            \includegraphics[width=\textwidth]{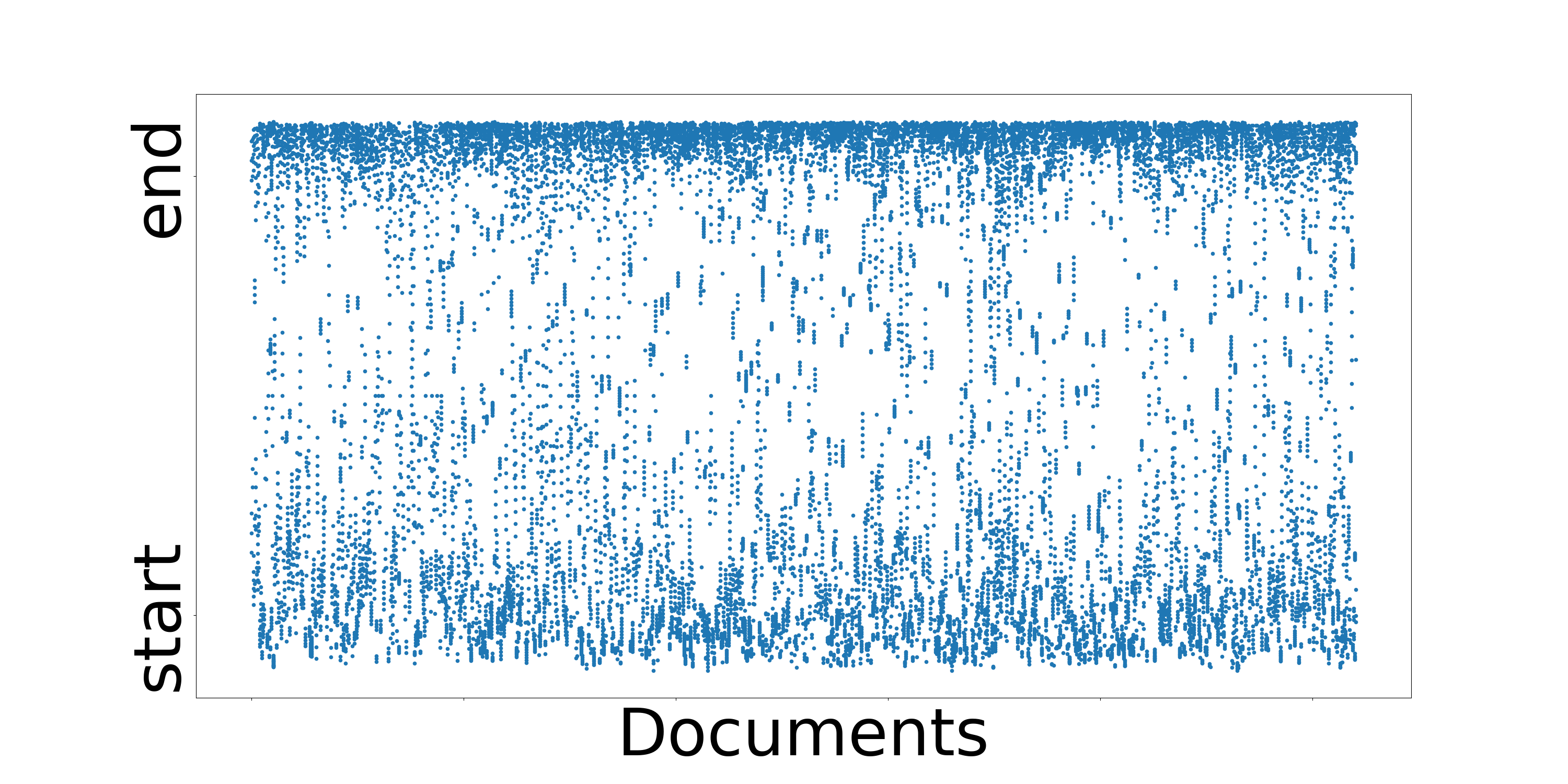}
            \caption[]%
            {{\small HipoRank without headers}}    
            \label{IRCoracle}
        \end{subfigure}
        \hfill
        \begin{subfigure}[b]{0.45\textwidth}   
            \centering 
            \includegraphics[width=\textwidth]{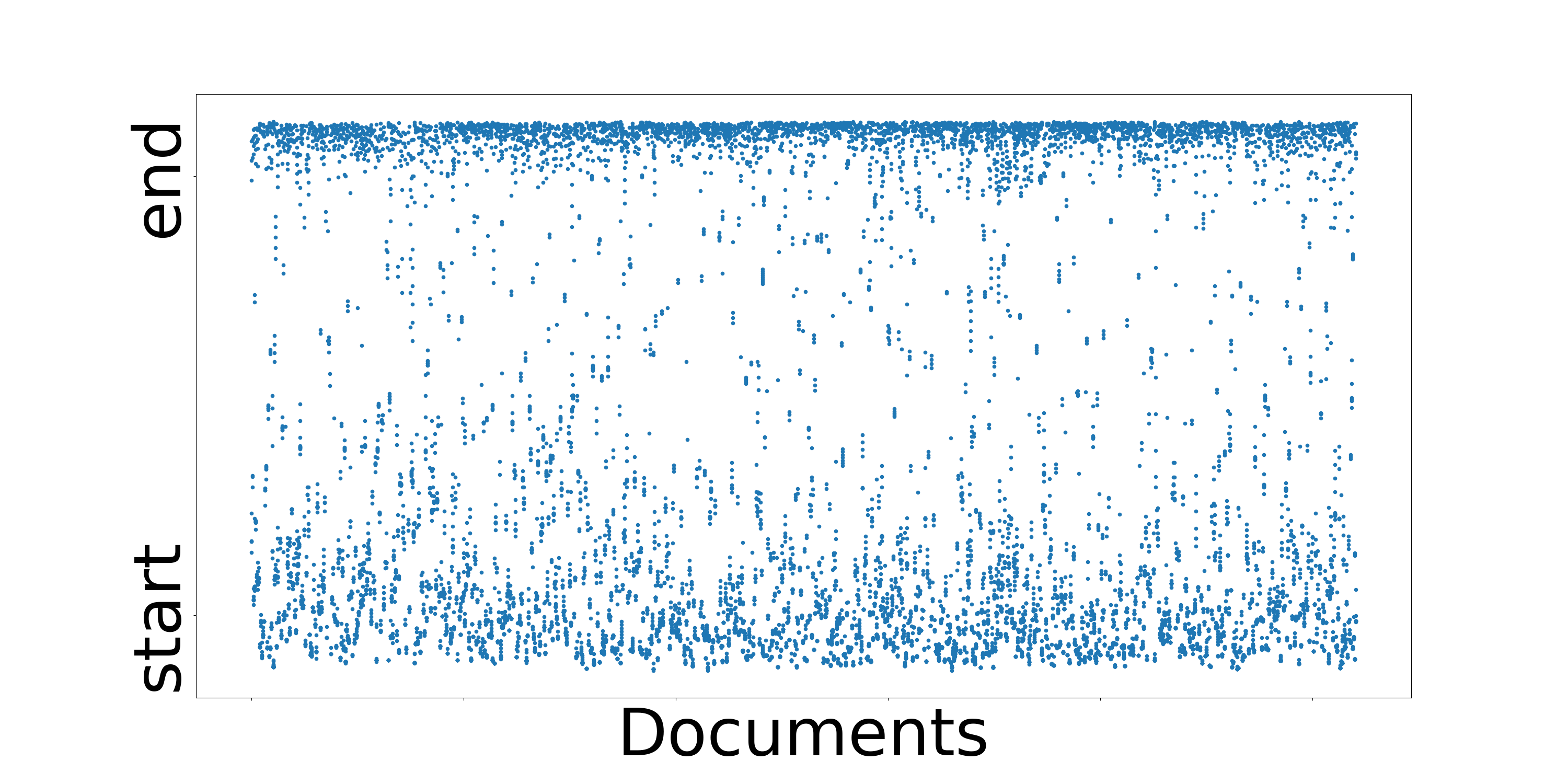}
            \caption[]%
            {{\small HipoRank without headers and with reweighting}}    
            \label{fig:mean and std of net34}
        \end{subfigure}
        \vskip\baselineskip
        
        \caption[]
        {\small Sentence positions in source cases for extractive summaries generated by different models using the original document structure on the
 test set. For (b) (c) (d), documents on the x-axis are
sorted in the same order. For IRC Oracles, \textcolor{red}{Issue}, \textcolor{reasonblue}{Reasoning} and \textcolor{conclusiongreen}{Conclusion} sentences are colored accordingly. } 
        \label{fig:comparison}
    \end{figure*}
    
    Finally, to better understand the behavior of different enhancements to the HipoRank backbone model, Figure \ref{fig:comparison} visualizes the positions of IRC sentences in the original article that are selected by a particular summarization method.  Plot (a) shows that the human-annotated IRC sentences in the summary tend to span across the source documents, with issues appearing in the beginning and conclusions in the end. Plot (b) shows that although HipoRank using the original document structure can successfully pick middle section sentences, the darkest band at the starting positions shows that the model still heavily relies on the inductive bias to pick the beginning sentences. Plot (c)
shows that removing the headers reduces the  starting sentence bias.  Finally, plot (d) shows that 
reweighting  reduces the number of sentences appearing on both ends. Further analyses on the complete automatic evaluation results\footnote{See Appendix \ref{appendix:reweight_effects} for ROUGE precision and recall.} suggest that the improvements come from higher recall values.

\subsection{Argumentative Sentence Coverage}\label{sec:salient_eval}
Taking advantage of the sentence-level IRC annotations, we propose recall metrics to better measure the summary quality from a legal argumentation perspective ({\bf RQ3}).  We compute the recall of ``IRC'' sentences extracted from the original case as source IRC coverage (src. IRC). 
We similarly compute the coverage of IRC sentences in the human-written summaries as target IRC coverage (tgt. IRC) and all sentences as target sentence coverage (trg. cov.). To do so we apply the oracle summarizer (Section \ref{sec:oracle}) to map the generated extractive summaries to the human-written abstractive summaries. 


We report these values for the IRC oracle, 
an unsupervised (LexRank) and supervised (BERT\_EXT) baseline, the  discourse-aware HipoRank with the original structure, and our best reweighting model using C99-topic segmentation. Table \ref{tab: salient_coverage} shows that our model obtains the highest target IRC recall and coverage, suggesting that the  summaries are more similar to the references with respect to 
the decision's argumentation. Another unsupervised model, LexRank, obtains the highest source IRC, but its off-the-shelf package requires a fixed sentence ratio selected from the source. 
This produced longer summaries than other approaches and thus captured more IRCs in the source. 

\begin{table}[t!]
\small
    \centering
    \renewcommand{\arraystretch}{1}
    \begin{tabular}{c|c|c|c}
    \toprule
  \textbf{Model} & \textbf{src. IRC} & \textbf{tgt. IRC} & \textbf{trg. cov.} \\
  \midrule 
  \multicolumn{3}{l}{\textit{Oracle}}\\
  \midrule
  IRC & 1 (0.00) & 0.918 (0.18)  &  0.820 (0.25) \\
  \midrule 
     \multicolumn{3}{l}{\textit{Baselines}}\\
  \midrule
  BERT\_EXT & 0.804 (0.27) & 0.846 (0.23) &  0.833 (0.23)\\
  LexRank & \textbf{0.912} (0.19) & 0.811 (0.26) &  0.800 (0.27)\\
  HipoRank & 0.800 (0.25) & 0.851 (0.24) & 0.844 (0.22) \\
   
  \midrule 
  {\it Ours} & 0.823 (0.26) & \textbf{0.866} (0.20) & \textbf{0.850} (0.21) \\
  \bottomrule

    \end{tabular}
    \caption{Average recall of IRC sentences matched in the original case (src. IRC), gold summary (tgt. IRC), as well as target sentences coverage (trg. cov.) for each document (standard deviation in parenthesis).}
    \label{tab: salient_coverage}
\end{table}

\subsection{Human Evaluation Discussion}


As a first step towards human evaluation, we tried to extend the HipoRank  setup in \citet{dong-etal-2021-discourse} and designed a human evaluation protocol as follows. We asked human judges\footnote{All judges should be native English speakers who are at least pursuing a JD degree in law school and have experience in understanding case law.} to read the human-written reference summary and presented  extracted sentences from different summarization systems. The judges were asked to evaluate a system-extracted sentence according to two criteria: (1) \textit{Content Coverage} -  whether the presented sentence contained content from the human summary, and (2) \textit{Importance} - whether the presented sentence was important for a goal-oriented reader even if it was not in the human summary\footnote{Here we assumed the goal-oriented reader as the lawyers or law students seeking information from the case.}. The sentence selection was anonymized and randomly shuffled. We used the same sampling strategy in \citet{dong-etal-2021-discourse} to pick ten reference summaries where the system outputs were neutral (i.e., had similar R-2 scores compared to the human reference). 
However, initial annotation on a small set by a legal expert demonstrated that the selected sentences may not reflect the model's capability. Most sampled system outputs had low ROUGE-2 F1 scores compared to the reference (normally below 10\% while the average model performance is 17\%), and the human evaluator reported that some of the selected sentences were not meaningful. We thus propose that a more careful sampling technique will be required for legal annotation tasks such as ours. 

To further guide our future work, we also reviewed how prior legal domain research has performed human evaluations when automatically summarizing legal documents  \cite{polsley-etal-2016-casesummarizer, zhong-2019-iterativemasking, salaun-2022-jurix}. Due to the burden of reading lengthy original documents, as in our human evaluation, most prior work evaluated summary quality using reference summaries rather than  source documents. In addition,  legal evaluations have typically been small-scale 
(5-20  summaries) due to the need to have evaluators with particular types of expertise (e.g., law graduate students or law professors), which was a similar constraint in our exploratory human evaluation.  Researchers have also designed new types of legally-relevant evaluation questions that  evaluate the summary for task-specific properties that go beyond more typical properties such as grammar, readability, and style. In our case, we would like legal experts to assess IRC coverage in the future.

\section{Conclusion}
We presented an unsupervised graph-based model
for the summarization of long legal case decisions. Our
proposed approach incorporated diverse views of the document structure of legal cases and utilized a reweighting scheme to better select argumentative sentences. Our exploration of document structure demonstrates how using different types of document structure impacts summarization performance. Moreover, a document structure inspired reweighting scheme yields  performance gain on the CanLII case dataset.

\section*{Ethical Considerations}
The utilization of the generated summary results of legal documents remains important. Current extractive methods avoid the problem of generating hallucinated information \cite{kryscinski-etal-2020-evaluating,maynez-etal-2020-faithfulness}, which has been observed in abstractive methods that use large-scale pre-trained language models. The extracted sentences, however, may not capture the important contents of the legal documents.
Meanwhile, CanLII has taken measures to limit the disclosure of defendants' identities (such as blocking search indexing). Thus, using the dataset may need to be taken good care of and avoid impacting those efforts. 

\section*{Acknowledgement}
We thank the Pitt AI Fairness and Law group,
the Pitt PETAL group, and the
anonymous reviewers for their valuable feedback.
This research is supported by
the National Science Foundation under Grant IIS-2040490 and a gift from Amazon.

\section*{Limitations}
The dataset we used has a relatively small scale (1K) 
test set. Meanwhile, the automatic evaluation metrics may fall short compared to human evaluations, 
thus unfaithfully representing the final quality of generated summaries. Although lightweight, there is still a large performance gap between our unsupervised method and both the extractive oracles as well as abstractive models (Appendix \ref{sec:appendix_abstractive}), especially given the small-scale training data. There are more graph-based methods to aggregate information from the built graphs and we would like to explore and include more graph-based methods but selected the most relevant one in this work.
Moreover, our proposed reweighting paradigm heavily relied on observations about the structure of legal cases. Many other legal document types, such as bills and statutes, have inherently distinct structures. Our results also show the importance of finding the correct  structure and  weights, which can vary depending on the corpus. This will require more advanced methods to find the correct structure and weights for a dataset.




\bibliography{custom}
\bibliographystyle{acl_natbib}

\appendix

\newpage
\clearpage
\section{The HipoRank Algorithm}\label{sec:hiporank}
In this section, we provide a detailed recap  of the HipoRank algorithm \cite{dong-etal-2021-discourse}. Our approach mainly modifies the obtained document graphs by building a \textit{section-section} graph and changes the final summary selection algorithms.  
\subsection{Hierarchical Document Graph Creation}
\begin{figure}[h]
\centering
 \includegraphics[width=\linewidth]{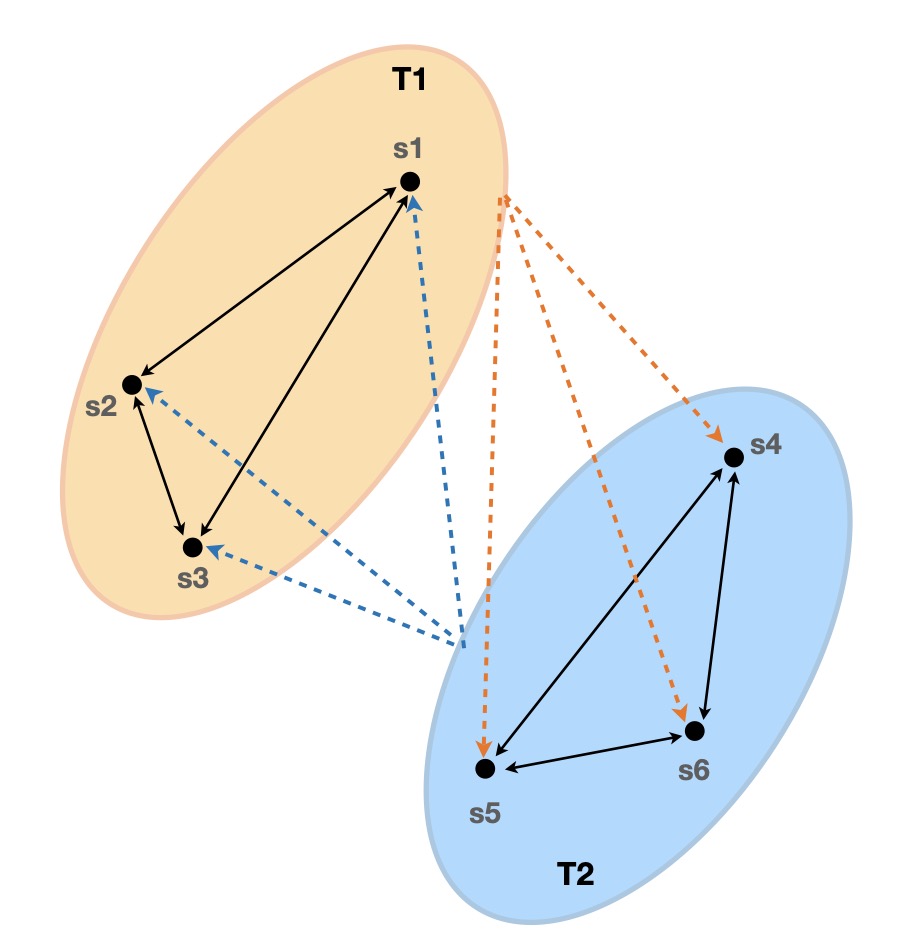}
  \caption{ (Reproduced from \cite{dong-etal-2021-discourse}) An example of a hierarchical document graph 
constructed by HipoRank approach on a toy document, which
contains two sections \{T1, T2\}, each containing three
sentences for a total of six sentences \{s1, . . . , s6\}. In the graph, 
each double-headed arrow represents two edges with
opposite directions. The solid and dashed arrows indicate intra-section and inter-section connections respectively.}
  \label{fig:hiporank_example}
\end{figure}

The document is first split into its sections, then into sentences. Two levels of connections are allowed in the built hierarchical graph: intra-sectional connections and inter-sectional connections. Following the original paper, we display a toy example of these two types of connections in Figure \ref{fig:hiporank_example}.

\paragraph{Intra-sectional connections} are designed to measure a sentence's importance score inside its section. The authors built a fully-connected subgraph over all sentences in the same section, allowing for \textit{sentence-sentence} edges, which are measured by a weighted version of the similarities of sentence embeddings.  
\paragraph{Inter-sectional connections} ``aim to model the
global importance of a sentence with respect to
other topics/sections in the document'', according to \citet{dong-etal-2021-discourse}. To reduce the expensive computation of all sentence-sentence connections spanning across a document, HipoRank's authors introduce section nodes on top of sentence nodes, and only allow for \textit{sentence-section} edges to model the global information. 
\subsection{Asymmetric Edge Weighting by
Boundary Functions}
In order to compute the weight of an edge, HipoRank measures the similarity of sentence-sentence pairs by computing the cosine similarity of encoded sentence embeddings. Similarly, for sentence-section pairs, it averages the sentences' representations in the same section, uses it as the section vector, and further computes the cosine similarity. Taking two discourse hypotheses of long scientific documents into account ((1) important sentences are near the
boundaries (start or end) of a text \cite{5392648} and (2) sections near the text boundaries (start or end) are more important \cite{teufel-1997-sentence}), the authors of HipoRank capture this asymmetry by making their hierarchical graph directed and inject asymmetric edge
weighting over intra-section and inter-section connections. We refer to the original paper for more detailed setups and algorithm details.
\subsection{Importance Computation and Summary Generation}
We talk about the importance computation approach and summary generation details in \S \ref{sec:main_paper_hiporank}. 

\section{Training Details and Hyperparameters}\label{appendix:detail}
All of our experiments are conducted on Quadro RTX 5000 GPUs, each of which has 16 GB RAM. For the extractive oracle baseline, we use the python package of rouge\footnote{\url{ https://pypi.org/project/rouge/}} to compute the ROUGE-L scores for sentence scoring. \paragraph{Document Segmentation}
We provide details of segmentation methods mentioned in \S \ref{sec:views} below. For sentence encoding, we use the sentence\_transformer library\footnote{\url{https://www.sbert.net/}}, and the checkpoint of ``bert-base-nli-stsb-mean-tokens'' for sentence representations. For the HMM stage segmentation, we train a GaussianHMM model with hmmlearn\footnote{\url{https://hmmlearn.readthedocs.io/en/latest/}}, setting the number of the components at 5 and train the model for 50 iterations on the validation set. For C99 algorithm, we use an implementation\footnote{\url{https://github.com/GT-SALT/Multi-View-Seq2Seq/blob/master/data/C99.py}} shared from \citet{chen-yang-2020-multi} in their original paper. We set the window size of 4 and std\_coefficient as 1. All data processing scripts are publicly available in a combined package in \url{https://github.com/cs329yangzhong/DocumentStructureLegalSum}. 

\paragraph{Supervised Model} We build our BERT\_EXT, the extractive model, on top of the official code base of the work of \citet{liu-lapata-2019-text}\footnote{\url{https://github.com/nlpyang/PreSumm}}. Since many original documents' lengths  go beyond the 512 token limits, we break the full document into different chunks and train the model to extract the top-3 sentences. For hyperparameters, we use 4 GPUs, set the learning rate of 2e-3, and save the best checkpoints at every 5,000 steps. We set the batch size as 3,000, the maximum training step at 100,000, and warm-up steps at 10,000.

\paragraph{Unsupervised Models} 
We use off-the-shelf packages for most traditional models. We use LSA\footnote{\url{https://github.com/luisfredgs/LSA-Text-Summarization}}, TextRank\footnote{\url{https://github.com/summanlp/textrank}}, and LexRank\footnote{\url{https://github.com/crabcamp/lexrank}} accordingly.

For PACSUM model, we follow the re-implementation\footnote{\url{https://github.com/mirandrom/HipoRank}} of \cite{dong-etal-2021-discourse} and keep the hyperparameters fixed with the original setup. BERT-based sentence embeddings are extracted using the fine-tuned BERT model released from the original paper \cite{zheng-lapata-2019-sentence}. We also experimented with LEGAL-BERT \cite{chalkidis-etal-2020-legal} in the early stages of our research but found it degraded performance on the baselines. 

For HipoRank, we use the publicly available code base\footnote{\url{https://github.com/mirandrom/HipoRank}}.  We experimented with various hyperparameter settings on the validation set but we find that
the original hyperparamters used in the original paper for PubMed dataset seem to be the most
stable and produce the best results. ($\lambda_1 = 0.0$, $\lambda_2 = 1.0$, $\alpha = 1.0 $, with $\mu_1 = 0.5$.)

We build our reweighting model on top of the HipoRank dataset. We search the threshold g (for phase transition between phases 1 and 2) between [0.3, 0.5, and 0.7], finding that 0.5 is the best for the CanLII dataset.





\section{The Effects of Reweighting Algorithm}\label{appendix:reweight_effects}
We  study the effects of our reweighting algorithm by comparing different models' performances on the input documents with original structures. As shown in Table \ref{tab:original_structure_p_r_f1}, with a minor sacrifice of precision, the recall values are greatly improved with the reweighting algorithm, thus resulting in the final improvements of F1 scores.

\begin{table*}[]
    \centering
    \begin{tabular}{c|ccc|ccc|ccc}
    \toprule
   Document Structures & \multicolumn{3}{c}{ROUGE-1} & \multicolumn{3}{c}{ROUGE-2} & \multicolumn{3}{c}{ROUGE-L} \\
    \midrule 
       & P & R & F1 & P & R & F1 &P & R & F1  \\
       \midrule
       w/o header  &  45.24 & 47.39 & 42.58 & 19.23 & 20.12 & 18.01 & 42.25 & 43.95 & 39.63 \\
       \midrule 
       w/o header + Reweighting & 44.13 & 49.80 & 43.14 & 18.97 & 21.35 & 18.46 & 41.29 & 46.26 & 40.23 \\
      \bottomrule 
    \end{tabular}
    \caption{The Precision (P), Recall (R) and F1 of ROUGE-1/2/L scores for the inputs with original document structures, with and without reweighting algorithm. We find that the reweighting algorithm improves the recall, suggesting that more argumentative sentences in the references are covered. }
    \label{tab:original_structure_p_r_f1}
\end{table*}


\section{Examples}\label{appendix:1}

\subsection{Summary Generation Results}\label{appendix:output}
We show the reference, best baseline, and our model's output on the C99-topic view of the without header version of documents in Table \ref{tab:canlii_example}.
\begin{table*}
\small
\begin{tabular}{l|l}
\toprule
    Model &  Summary \\
    \midrule
    \multirow{10}{*}{Reference} & FIAT: The defendants, Sims, Garbriel and Dumurs, bring separate motions, pursuant to \\
    & Queen's Bench Rule 41(a), for severance of the claims against them or for an order \\
    & staying the claims against them until the plaintiffs' claim against the primary defendant,  \\
    & Walbaum have been heard and decided. || HELD: 1) || The Court will look at all of the \\
    & circumstances in deciding whether to grant an application for severance. || In this case \\
    & the plaintiffs should not be precluded from adducing evidence related to Walbaum's \\
    & dealings with each of the applicants or required to segregate the evidence into two, \\
    & three or four separate trials. || Given the likelihood that the applicants will be required \\
    & to attend portions of the trial in respect of the Walbaum Group in any event, severance \\
    & would not necessarily result in a significant saving of time and expense. || 2) The plain- \\
    &-tiffs acknowledge that only relatively small portion of trial time (perhaps less than 1 \\
    & day) will pertain to the claims against any one of the Sims, Gabriel or Dumurs. || It wou-  \\
    & -ld be unfair to require all of the applicants to participate in all of the trial when very little \\
    & of it will be relevant to them. || Specific dates and times should be set aside for the \\
    & plaintiff to call evidence with respect to its claims against each applicant group. || The \\
    & applicants should be relieved from attending the trial at any other time. \\
    \midrule
    \multirow{13}{*}{HipoRank} & QUEEN’S BENCH FOR SASKATCHEWAN ||
2007 SKQB 296 || Judicial Centre: Regina \\
&|| DUN-RITE PLUMBING \& HEATING LTD. || (d) Robert Dumur || 593340 Saskatchewan \\
& Ltd., carrying on || business as Dumur Industries (herein “the Dumurs”) || [2] The Sims, Gab-\\
& -riel and the Dumurs bring separate motions, pursuant to Queen’s Bench Rule 41(a) for \\
& severance of the claims against them or for orders staying the claims until the plaintiffs’ claims \\ 
& against the Walbaum Group have
been heard and decided. || ANALYSIS || [12] T applications \\
& are brought pursuant to Queen’s Bench Rule 41 which states: || 41 (1) Where the joinder of \\
& multiple claims or parties in the same action may unduly complicate or delay the trial, or cause \\
& undue prejudice to a party, the court may: || (a) order separate trials; || [21] It will also be left to \\
& the trial judge (or the pre-trial management judge) to designate specific days on which defence \\
& evidence may be adduced during the trial and argument presented with respect to each claim. || \\
& All defendants other than those comprising the Walbaum Group shall be relieved from attending \\
& the trial on any date not designated by counsel for the plaintiffs or designated by the trial judge \\
& for adducing defence evidence and presenting argument. || [22] Costs will be in the cause.
|| D.P. \\
& Ball \\
    \midrule
    \multirow{10}{*}{Ours} & FIAT BALL J. || August 14, 2007 || [1] The plaintiff brings this action against nine  defendants \\
    & (the claim against the defendant Albert Fazakas has been discontinued) who can be separated \\
    & into four groups: 
    || All-Rite Plumbing Heating Ltd. || [18] Although choeunsel for the plaintiffs \\
    & asserts that the evidence against all of the defendants can be adduced in no more than two and \\
    & one-half days, given the number and complexity of the claims against the Walbaum Group this \\
    & estimate seems very unrealistic. || [19] The plaintiffs acknowledge that only relatively small \\
    & portion of trial time (perhaps less than one day) will pertain to the claims against any one of the \\
    & Sims, Gabriel or the Dumurs. || It would be unfair to require all of the applicants to participate \\
    & in all of the trial when very little of it will be relevant to them. || The applicants should be \\
    & relieved from attending the trial at any other time. || The plaintiffs shall not call evidence in  \\
    & respect of those claims on any other date without leave of the court. || All defendants other than \\
    & those comprising the Walbaum Group shall be relieved from attending the trial on any date not \\
    & designated by counsel for the plaintiffs or designated by the trial judge for adducing defence \\
    & evidence and presenting argument. \\
\bottomrule
\end{tabular}
\caption{Generated summaries for a CanLII case decision (ID: 2\_2007skqb296), we use special symbol ``||'' to mark the sentence boundaries.}
\label{tab:canlii_example}
\end{table*}

 \subsection{IRC Annotation}\label{appendix:irc}
\begin{figure*}
\centering
 \includegraphics[width=\linewidth]{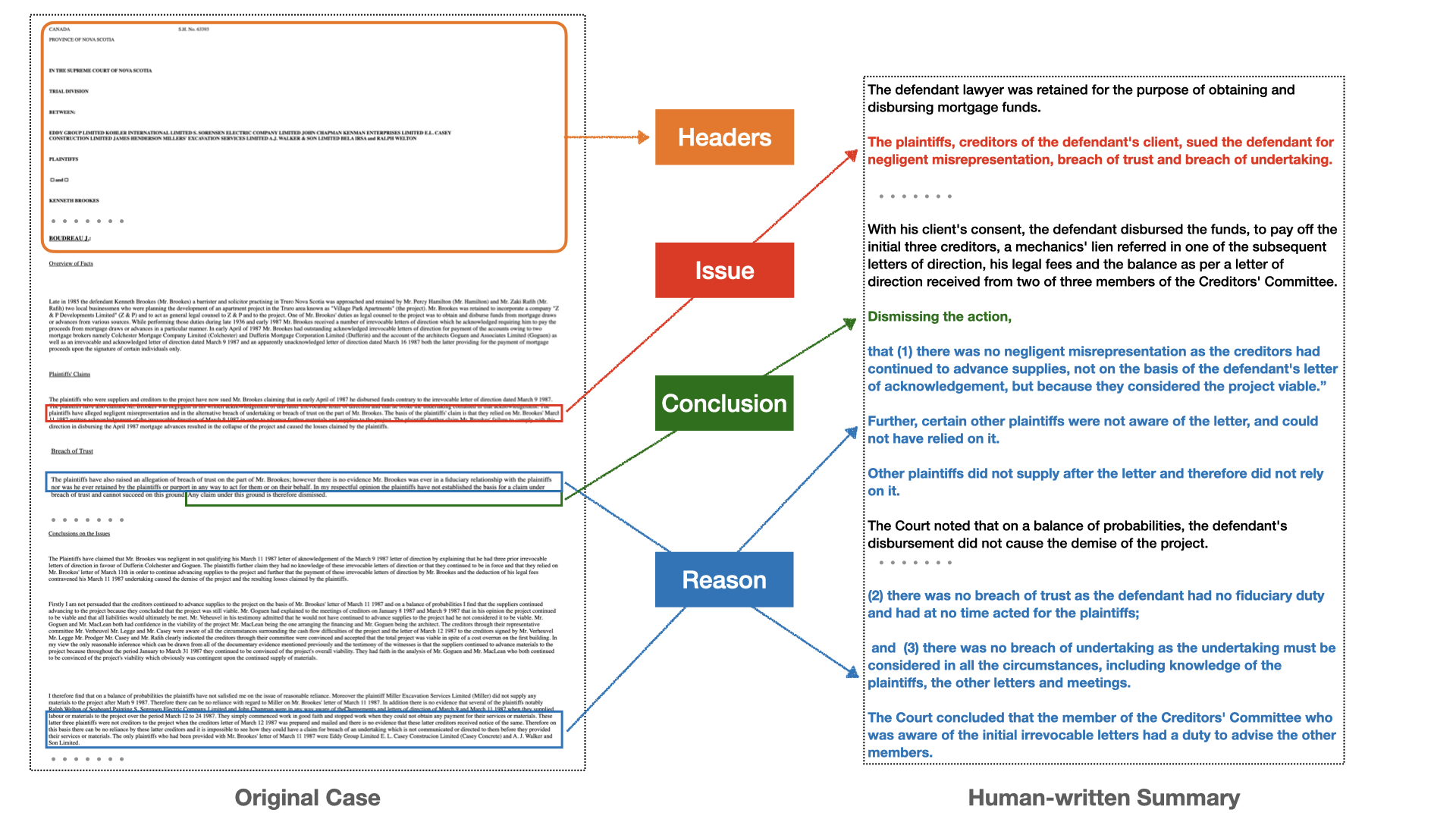}

  \caption{An example of the annotated \textcolor{red}{Issue}, \textcolor{cyan}{Reason}, and \textcolor{teal}{Conclusion} sentences in CanLII dataset's case and summary pair (ID: 1991canlii4370). A portion of the beginning sentences in the case are not as important as the main document, including the meta-data of the case such as the participants' names, time, locations, etc. Thus, we treated them as headers and filtered them out using a heuristic introduced in Appendix \ref{sec:appendix_heuristics}. }
  \label{fig:IRC_example}
\end{figure*}

We show the IRC annotation of both a case and its human summary in Figure \ref{fig:IRC_example}.

\subsection{Document Cleaning Heuristics}\label{sec:appendix_heuristics}
The heuristics for filtering the headers from cases are provided below for replication purposes; we also provide the code\footnote{\url{https://github.com/cs329yangzhong/DocumentStructureLegalSum}} to process the CanLII data (although it requires that the data must first be obtained through an agreement with the Canadian Legal Information Institute). 

\begin{enumerate}
    \item Cut the document until the sentence begins with ``Introduction''. 
    \item Cut the document until the sentence starts with an ordered number such as (1), [1].
    \item Remove rows until the judge's name or case date appeared.
\end{enumerate}

\subsection{Comparing to Abstractive Summarization}\label{sec:appendix_abstractive}
For supervised abstractive baselines, we experiment with BART \cite{lewis-etal-2020-bart} and  Longformer-Encoder-Decoder (LED) \cite{Beltagy2020Longformer}. The latter model can process longer input documents up to 16k tokens. The results  in Table \ref{tab:unsupervised_results_abstractive} show that there still exists a gap between the extractive and abstractive models.

\begin{table*}[h!]
\small
    \centering
    \renewcommand{\arraystretch}{1}
    \begin{tabular}{c|l|cc}
    \toprule
        & & \multicolumn{2}{c}{\textbf{CanLII}}\\
      ID & Model  &  R1/R2/RL  & BS\\
         \midrule
         \multicolumn{3}{c}{{Oracles}} \\
         \midrule
          1 & IRC & 58.04/36.02/55.28 & 87.94 \\
        2 &  EXT & 59.38/38.77/56.94 & 87.85 \\

    \midrule 
    \multicolumn{3}{c}{{Supervised Extractive }} \\
    \midrule
    3 & BERT\_extractor & {43.44}/{17.84}/{40.36} & {84.47} \\
    \midrule
    \multicolumn{3}{c}{{Supervised Abtractive }} \\
    \midrule
   
    4 & BART & 50.50/25.58/46.82 &  87.25\\
    5 & LED & 53.72/28.75/ 50.17 & 87.55\\
    
    \bottomrule
    \end{tabular}
    \caption{The automatic evaluation results on the CanLII  test set with supervised abstractive models.}
    \label{tab:unsupervised_results_abstractive}
\end{table*}

\end{document}